\def\year{2019}\relax
\PassOptionsToPackage{dvipsnames,table,xcdraw}{xcolor}
\documentclass[letterpaper]
{article} 
\usepackage{aaai19}  
\usepackage{times}  
\usepackage{helvet}  
\usepackage{courier}  
\usepackage{url}  
\usepackage{graphicx}  
\usepackage{xspace}
\usepackage{rotating}
\usepackage{longtable}
\usepackage{capt-of}
\usepackage{color}
\usepackage{textgreek}
\usepackage{multirow}
\usepackage[normalem]{ulem}
\usepackage[algo2e,ruled,vlined,linesnumbered]{algorithm2e}
\usepackage{amsmath}
\usepackage{subfigure}
\usepackage{tabulary,booktabs}

\newcommand{\rn}{robot network\xspace}
\newcommand{\tn}{target network\xspace}

\newcommand{\hp}{hyperparameter\xspace}
\newcommand{\hps}{hyperparameters\xspace}

\newcommand{\monas}{MONAS\xspace}
\newcommand{\monass}{MONAS-S\xspace}

\newcommand{\baselineB}{CondenseNet\xspace}
\newcommand{\alexnet}{AlexNet\xspace}
\newcommand{\condensenet}{CondenseNet\xspace}

\newcommand{\hide}[1]{}

\title{MONAS: Multi-Objective Neural Architecture Search}

\begin{document}

\author{\AND \large Chi-Hung Hsu,\textsuperscript{1}
Shu-Huan Chang,\textsuperscript{1}
Jhao-Hong Liang,\textsuperscript{1}
Hsin-Ping Chou,\textsuperscript{1}\AND
\large Chun-Hao Liu,\textsuperscript{1}
Shih-Chieh Chang,\textsuperscript{1}
Jia-Yu Pan,\textsuperscript{2}
Yu-Ting Chen,\textsuperscript{2}\AND
\large Wei Wei,\textsuperscript{2}
Da-Cheng Juan\textsuperscript{2}\\\\
\textsuperscript{1}{National Tsing-Hua University, Hsinchu, Taiwan}
\textsuperscript{2}{Google, Mountain View, CA, USA}\\
\{charles1994608, tommy610240, jayveeliang, alan.durant.chou, newgod1992\}@gmail.com,\\
scchang@cs.nthu.edu.tw, 
\{jypan, yutingchen, wewei, dacheng\}@google.com}

\maketitle

\begin{abstract}
Recent studies on neural architecture search have shown that automatically designed neural networks perform as good as expert-crafted architectures. While most existing works aim at finding architectures that optimize the prediction accuracy, these architectures may have complexity and is therefore not suitable being deployed on certain computing environment (e.g., with limited power budgets). We propose \monas, a framework for \underline{M}ulti-\underline{O}bjective \underline{N}eural \underline{A}rchitectural \underline{S}earch that employs reward functions considering both prediction accuracy and other important objectives (e.g., power consumption) when searching for neural network architectures. Experimental results showed that, compared to the state-of-the-arts, models found by \monas achieve comparable or better classification accuracy on computer vision applications, while satisfying the additional objectives such as peak power.

\end{abstract}

\section{Introduction}
\label{sec:intro}




Convolutional Neural Networks (CNN) have shown impressive successes in computer-vision applications; however, designing effective neural networks heavily relies on experience and expertise, and can be very time-consuming and labor-intensive.
To address this issue, the concept of neural architecture search (NAS) has been proposed~\cite{NASWITHRL}, and several literatures have shown that architectures discovered from NAS outperforms state-of-the-art CNN models on prediction accuracy~\cite{baker2016designing,NASWITHRL}. 
However, models solely optimized for prediction accuracy generally have high complexity and therefore may not be suitable to be deployed on computing environment with limited resources (e.g., battery-powered cellphones with low amount of DRAM).

In this paper, we propose \monas, a framework for Multi-Objective Neural Architectural Search that has the following properties:
\begin{itemize}
\item \textbf{[Multi-Objective]} \monas considers both prediction accuracy and additional objectives (e.g., energy consumption) when exploring and searching for neural architectures.

\item \textbf{[Adaptability]} \monas allows users to incorporate customized constraints introduced by applications or platforms; these constraints are converted into objectives and used in \monas to guide the search process. 

\item \textbf{[Generality]} \monas is a general framework that can be used to search a wide-spectrum of neural architectures. In this paper, we demonstrate this property by applying \monas to search over several families of CNN models from \alexnet-like~\cite{krizhevsky2012imagenet},  \condensenet-like~\cite{huang2017condensenet}, to ResNet-like~\cite{Kaiming2015} models.

\item \textbf{[Effectiveness]} Models found by \monas achieves higher accuracy and lower energy consumption, which outperforms the state-of-the-art CondenseNet~\cite{huang2017condensenet} in both aspects. Experimental results also confirm that \monas effective guides the search process to find models satisfying the predefined constraints. 

\item \textbf{[Scalability]}
To make \monas more scalable, we further extended \monas into \monass to accelerate the search process by adopting the weight-sharing searching technique\footnote{The weight-sharing searching technique is first proposed by Pham et al.~\cite{enas}}. Compared to \monas, \monass is able to search for a design space up to $10^{22}$ larger, and at the same time, \monass is 3X faster.


\end{itemize}

The remainder of this paper is organized as follows. Section 2 reviews the previous work of NAS. Section 3 details the proposed \monas and \monass. Section 4 describes the experimental setup. Section 5 provides the experimental results and Section 6 concludes this paper.

\section{Background}
\label{sec:background}

Recently, automating neural network design and hyperparameter optimization has been proposed as a reinforcement learning problem. Neural Architecture Search \cite{NASWITHRL,2017arXiv170707012Z} used a recurrent neural network (RNN) to generate the model descriptions of neural networks and trained this RNN with reinforcement learning to maximize the validation accuracy of the generated models.  Most of these works focus on discovering models optimized for high classification accuracy without penalizing excessive resource consumption.

Recent work has explored the problem of architecture search with multiple objectives. Pareto-NASH~\cite{2018arXiv180409081E} and Nemo~\cite{NEMO} used evolutionary algorithms to search under multiple objectives. \cite{sigopt} used Bayesian optimization to optimize the trade-off between model accuracy and model inference time.


The major difference between the previous works and ours is that we consider the model computation cost, such as power consumption or \underline{M}ultiply-\underline{AC}cumulate (MAC) operations, as another constraint for architecture search. In this paper, our optimization objectives include both CNN model accuracy and its computation cost. We explore the model space by taking the peak power or accuracy to be a prerequisite, or we can directly set our objectives weights when considering the trade-off between model accuracy and computation cost.


\section{Proposed Framework: \monas}
\label{sec:method}
In this section, we describe our \monas framework for neural architecture search with \textit{multiple objectives}, which is built on top of the framework of reinforcement learning.

\subsection {Framework Overview}
Our method \monas adopts a two-stage framework similar to NAS \cite{NASWITHRL}.
In the generation stage, we use a RNN as a \textbf{\textit{\rn}} (\textit{RN}), which generates a \hp sequence for a CNN. In the evaluation stage, we train an existing CNN model as a \textbf{\textit{\tn}} (\textit{TN}) with the \hps output by the RNN. The accuracy and energy consumption of the \tn are the \textbf{rewards} to the \rn.  The \rn updates itself based on this reward with reinforcement learning.  Algorithm~\ref{alg:monas} gives the pseudo code of the overall procedure of our \monas framework.

\begin{algorithm2e}
\caption{\monas Search Algorithm for Neural Architectures.}
\label{alg:monas}
\KwIn{$Search Space$, Reward Function(R), $n_{iterations}$, $n_{epochs}$}
\KwOut{Current Best Target Network($TN^*$)}
$Reward_{max} \gets 0$\;
Initialize the robot network $RN$\;
\For{$i \gets 1$ \textbf{to} $n_{iterations}$} {
  $TN_i \gets RN.generateTN(SearchSpace)$\;
  Train $TN_i$ for $n_{epochs}$\;
  $Reward_i \gets R(TN_i)$\;
  Update $RN$ with $Reward_i$ with the policy gradient method\;
  \If{$Reward_i > Reward_{max} $} {
      $Reward_{max} \gets Reward_i $\;
      $TN^* \gets TN_i$\;
  }
}
\Return{$TN^*$}\;
\end{algorithm2e}



\subsection {Implementation Details}
\paragraph{Robot Network}
Fig.~\ref{fig:rnnflow} illustrates the workflow of our \rn, an RNN model with one-layer Long Short-Term Memory (LSTM) structure. At each time step, we predict the hyperparameters in Table~\ref{tab:hparams} one at a time. The input sequence of the LSTM is a vector whose length is the number of all \hp candidates (for example, 4 possible values for the number of filters, filter height and filter width will give a vector of length 4*3 = 12), and it is initialized to zeros. The output of the LSTM is fed into a softmax layer, and a prediction is then completed by sampling through the probabilities representing for each hyperparameter selection. For the input sequence of the next time step, we make a one-hot encoding at the position of the current selection we take. 

\paragraph{Target Networks \& Search Space}
We demonstrate the generality of \monas by applying to two families of CNN models. The first one, which we called \textit{AlexNet}, is the simplified version of the AlexNet \cite{krizhevsky2012imagenet} as described in the TensorFlow tutorial\footnote{www.tensorflow.org\//tutorials\//deep\_cnn}. For the two convolutional layers in our target AlexNet, our \rn selects a value in [8, 16, 32, 48, 64] for the number of filters; [3, 5, 7, 9] for filer height and filter width. The other one is the CondenseNet \cite{huang2017condensenet}, an efficient version of DenseNet \cite{huang2017densely}. We predict the \textbf{stage} and \textbf{growth} for the 3 dense blocks in the \baselineB. Our \rn selects a stage value in [6, 8, 10, 12, 14] and a growth value in [4, 8, 16, 24, 32]. 

\begin{table}[h]
\begin{minipage}[h]{.5\textwidth}
\centering
\includegraphics[width=0.8\textwidth]{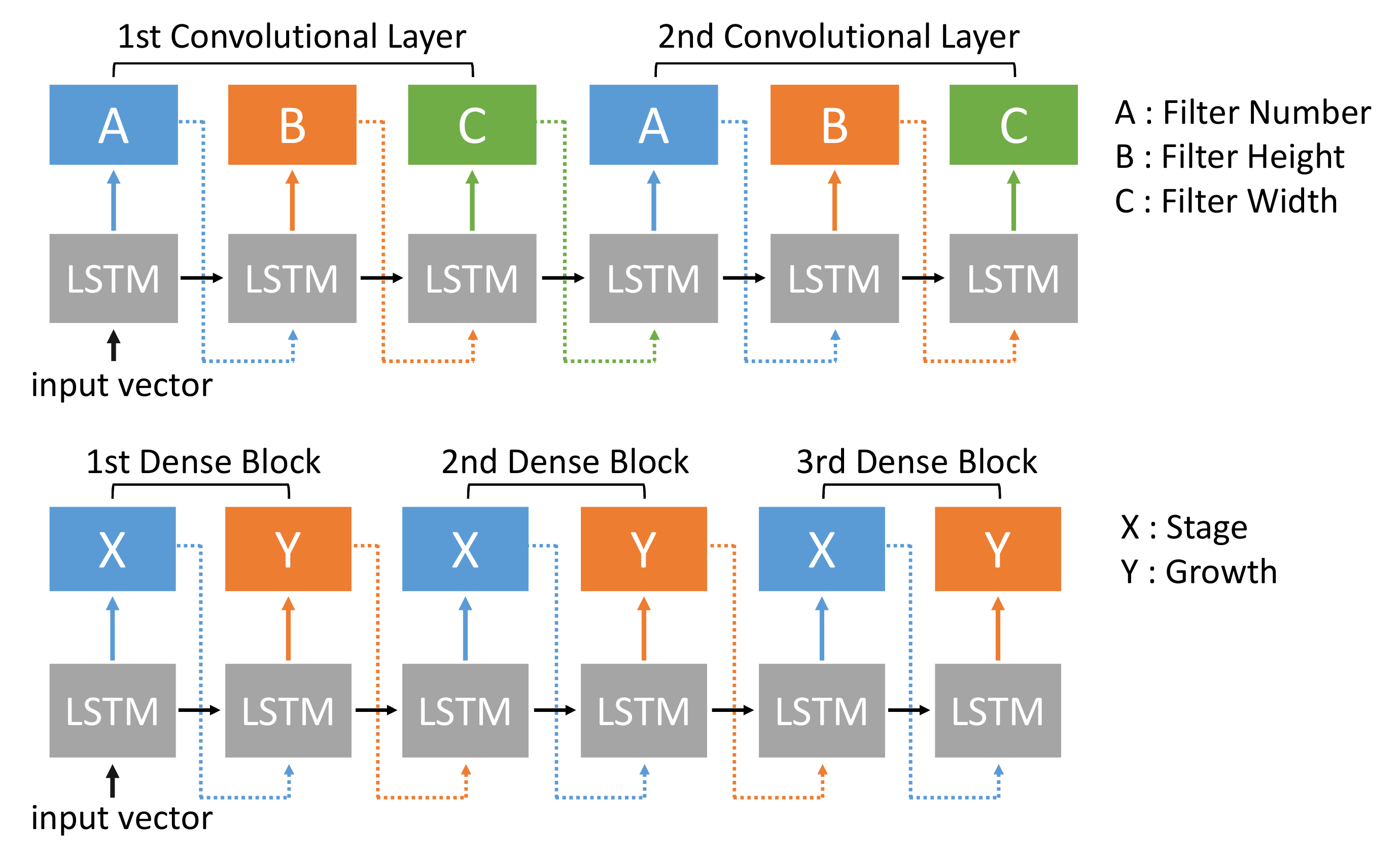}
\captionof{figure}{RNN workflow for \alexnet \newline and \condensenet}
\label{fig:rnnflow}
\vspace{12pt}
\end{minipage}
\begin{minipage}[h]{.5\textwidth}
\centering
\begin{tabular}{|c|c|} 
    \hline
      \textbf{Model} & \textbf{Hyperparameters} \\
      \hline 
      \multirow{3}{4em}{ AlexNet} & Number of Filters\\ 
                                 & Filter Height  \\
                                 & Filter Width  \\
      \hline
      \multirow{2}{6em}{CondenseNet} & Block Stage \\
                                     & Block Growth \\
      \hline
    \end{tabular}
    \caption{Hyperparameters used in experiments.}
    \label{tab:hparams}
\end{minipage}
\end{table}

\subsection {Reinforcement Learning Process}
In this subsection, we describe the details of our implementation of the policy gradient method to optimize the \rn. Then we describe the design of the reward functions.

\paragraph{Policy Gradient}
The hyperparameter decisions we make can be regarded as series of actions in the policy-based reinforcement learning \cite{policy}. We take the RNN model as an actor with parameter $\theta$, and $\theta$ is updated by the policy gradient method to maximize the expected reward $\bar{R}_{\theta}$:
\begin{equation}
\label{eq:update_theta}
\begin{aligned}
\theta_{i+1} = \theta_i + \eta\nabla\bar{R}_{\theta_i}
\end{aligned}
\end{equation}

where $\eta$ denotes the learning rate and $\nabla\bar{R}_{\theta_i}$ is the gradient of the expected reward with parameter $\theta_i$. We approximate the $\nabla\bar{R}_\theta$ with the method of \cite{NASWITHRL}:
\begin{equation}
\label{eq:gradient_R_bar}
\begin{aligned}
\nabla\bar{R}_\theta &= \sum_\tau\nabla P(\tau|\theta)R(\tau)\\&\approx \frac{1}{N}\sum_{n=1}^{N}\sum_{t=1}^{T}\nabla logP(a^n_t|a^n_{t-1:1},\theta)R(\tau^n)
\end{aligned}
\end{equation}



where $\tau$ = \{$a_1$, $a_2$..., $a_T$, $r_T$\} is the output from one \monas iteration. $R(\tau)$ is the reward of $\tau$, $P(\tau|\theta)$ is the conditional probability of outputting a $\tau$ under $\theta$. In our case, only the last action will generate a reward value $r_T$, which corresponds to the classification accuracy and power consumption of the \tn defined by a sequence of actions $a_{1:T}$. $N$ is the number of target networks in a mini-batch. By sampling $N$ times for multiple $\tau$s, we can estimate the expected reward $\bar{R}_\theta$ of the current $\theta$. In our experiments, we found that using $N$ = 1, which means the $\theta$ is updated every time for each \tn we sampled, improves the convergence rate of the learning process.


\paragraph{\monas for Scalability}
To improve the scalability and the training time for \monas, we adopt the techniques (such as weight sharing) in ENAS \cite{enas} with our \monas framework and propose the \monass method. 
Similar to ENAS, the training process of \monass is a two-step process: First, the shared weights of a directed acyclic graph (DAG) that encompasses all possible networks are pre-trained.  Then, the controller uses the REINFORCE policy learning method \cite{reinforce} to update the controller's weights -- models are sampled from the DAG, and their validation accuracy as evaluated on a mini-batch data is used as the reward of each model. Since \monass considers multiple objectives besides validation accuracy, other objective measures such as energy consumption are also computed for each of the sampled model, and together with the validation accuracy forms the reward to the controller.


\paragraph{Reward Function} The reward signal for \monas consists of multiple performance indexes: the validation accuracy, the peak power, and average energy cost during the inference of the target CNN model. To demonstrate different optimization goals, our reward functions are as follows:
\begin{itemize}
\item \textbf{Mixed Reward:} To find target network models having high accuracy and low energy cost, we directly trade off these two factors.
\begin{equation}
\label{eq:mix}
R = \alpha*Accuracy - (1-\alpha)*Energy
\end{equation}
\item \textbf{Power Constraint:} For the purpose of searching configurations that fit specific power budgets, the peak power during testing phase is a hard constraint here.
\begin{equation}
\label{eq:pow}
\begin{aligned}
R = Accuracy, \mbox{if power} < \mbox{threshold}
\end{aligned}
\end{equation}
\item \textbf{Accuracy Constraint:} By taking accuracy as a hard constraint, we can also force our RNN model to find high accuracy configurations.
\begin{equation}
\label{eq:accu}
\begin{aligned}
R = 1-Energy, \mbox{if accuracy} > \mbox{threshold}
\end{aligned}
\end{equation}
\end{itemize}

\begin{itemize}
\item \textbf{MAC Operations Constraint:} 
The amount of MAC operations is a proxy measure of the power used by a neural network. We use MAC in our experiments of \monass. By setting a hard constraint on MAC, the controller is able to sample model with less MAC.
\begin{equation}
\label{eq:mac_constraint}
R = Accuracy, \mbox{if MAC} < \mbox{threshold}
\end{equation}
\end {itemize}


In the equations~\eqref{eq:mix}\eqref{eq:pow}\eqref{eq:accu}, the accuracy and energy cost are both normalized to [0, 1]. For Eq~\eqref{eq:pow} and Eq~\eqref{eq:accu}, if current \tn doesn't satisfy the constraint, the reward will be given zero.
For Eq~\eqref{eq:mac_constraint}, a negative reward is given to the controller when the target network has a MAC that does not satisfy the given constraint.


\section{Experimental Setup}
\label{sec:exp}
\paragraph{Hardware:}
All the experiments are conducted via Python (version 3.4) with TensorFlow library (version 1.3), running on Intel XEON E5-2620v4 processor equipped with GeForce GTX1080Ti GPU cards.

\paragraph{GPU Profiler:}
To measure the peak power and energy consumption of the \tn for the reward to update our \rn, we use the NVIDIA profiling tool - \textbf{nvprof}, to obtain necessary GPU informations including peak power, average power, and the runtime of CUDA kernel functions. 


\paragraph{Dataset:} In this paper, we use the CIFAR-10 dataset to validate the \tn. We randomly select 5000 images from training set as the validation set and take the validation accuracy to be an objective of the reward function.






\paragraph{MAC Operations:}We calculate the MAC operations of a sampled target network based on the approach described in MnasNet \cite{2018arXiv180711626T}. We calculate MAC operations as follows:
\begin{itemize}
\item \textbf{Convolutional layer:}
\begin{equation}
\label{eq:con_mac}
K*K*C_{in}*H*W*C_{out}
\end{equation}
\item \textbf{Depthwise-separable convolutional layer:}
\begin{equation}
\label{eq:dep-con_mac}
C_{in}*H*W*(K*K+C_{out})
\end{equation}
\end{itemize}
In Eq~\eqref{eq:con_mac}~\eqref{eq:dep-con_mac}, $K$ is the kernel size of the filter, $H$/$W$ is the height/width of the output feature map and $C_{in}$, $C_{out}$ is the number of the input and output channels. We ignore the MAC operations of fully connected layer and skip connections since every target network has the same fully connected layer as the output layer and there is very few MAC operations in skip connections.

\section{Results and Discussions}
\label{sec:result}
In this section, we show some experimental results to illustrate the output of our proposed method.  We are in particular interested in the following questions: 


\begin{itemize}
\item Does \monas adapt to different reward functions?
\item How efficient does \monas guide the exploration in the solution space?
\item How does the Pareto Front change under different reward functions? 
\item Does \monas discover noteworthy architectures compared to the state-of-the-art?
\item Can \monas guide the search process while the search space is large?
\end{itemize}
\begin{figure}[h!]
\subfigure[Random Search]
{\includegraphics[width=0.48\textwidth]{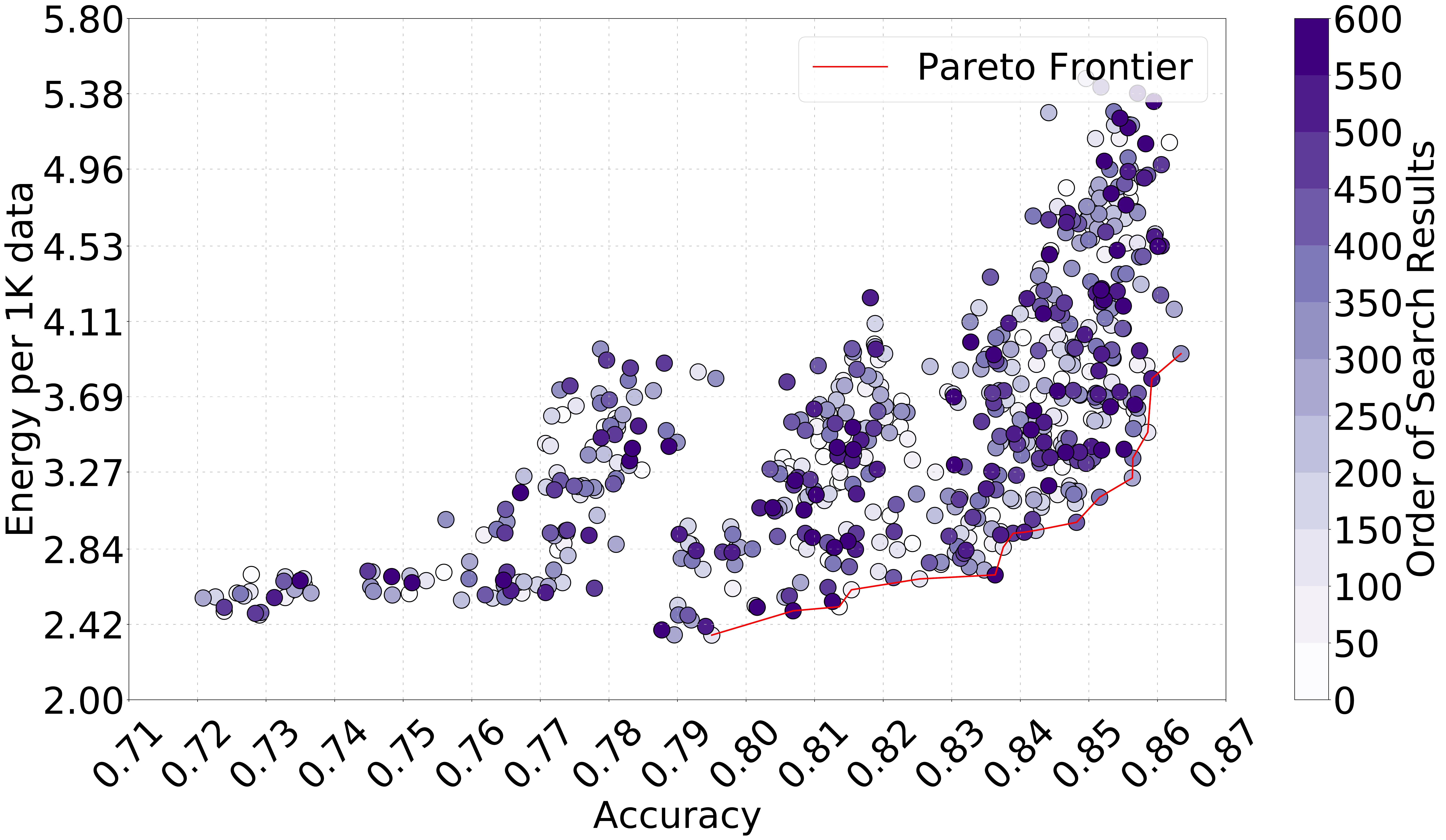}
\label{fig:random_alex_net}}
\subfigure[Max Power: 70W]
{\includegraphics[width=0.48\textwidth]{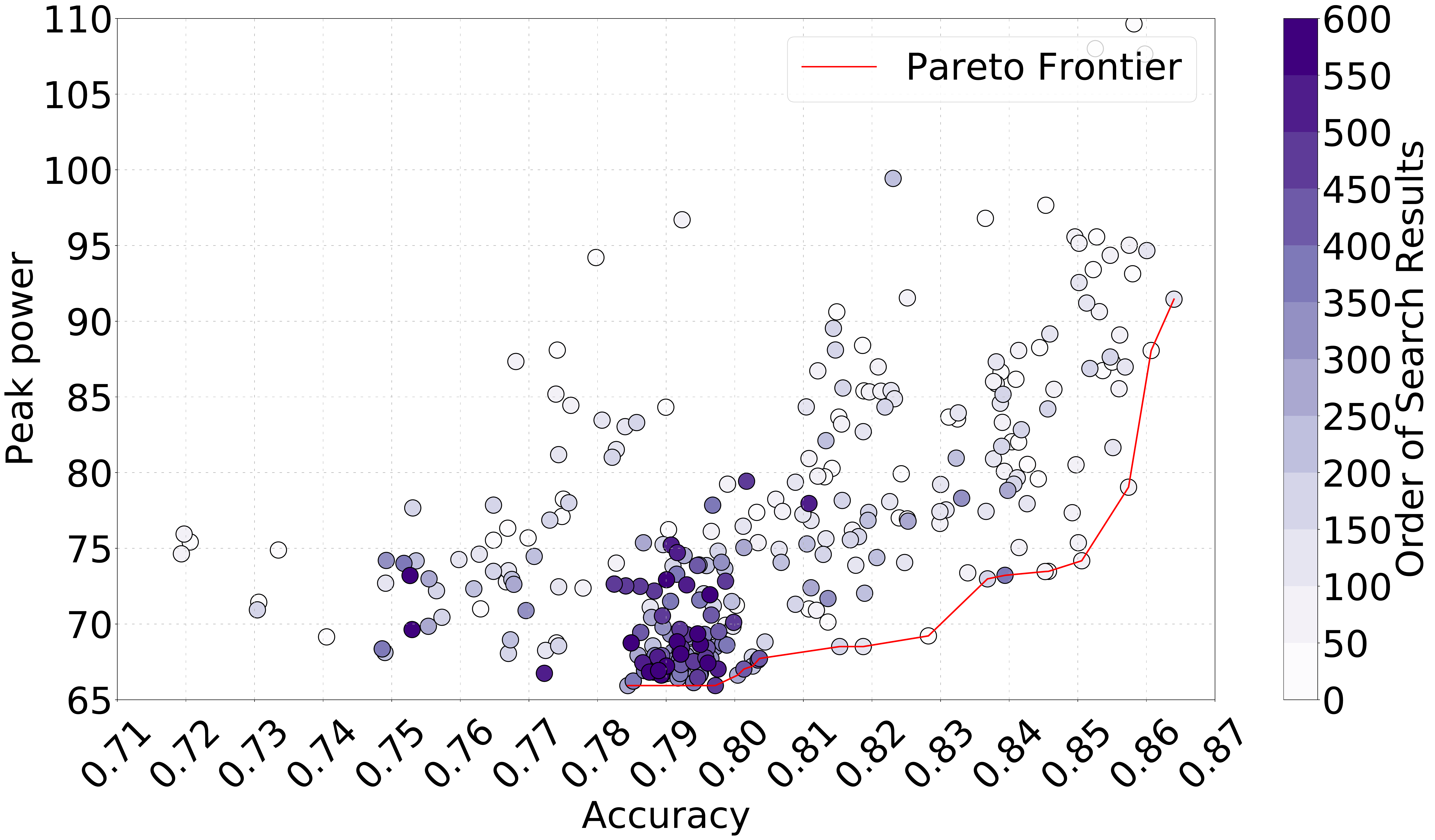}
\label{fig:70W_alex_net}}
\subfigure[Minimum Accuracy: 0.85]
{\includegraphics[width=0.48\textwidth]{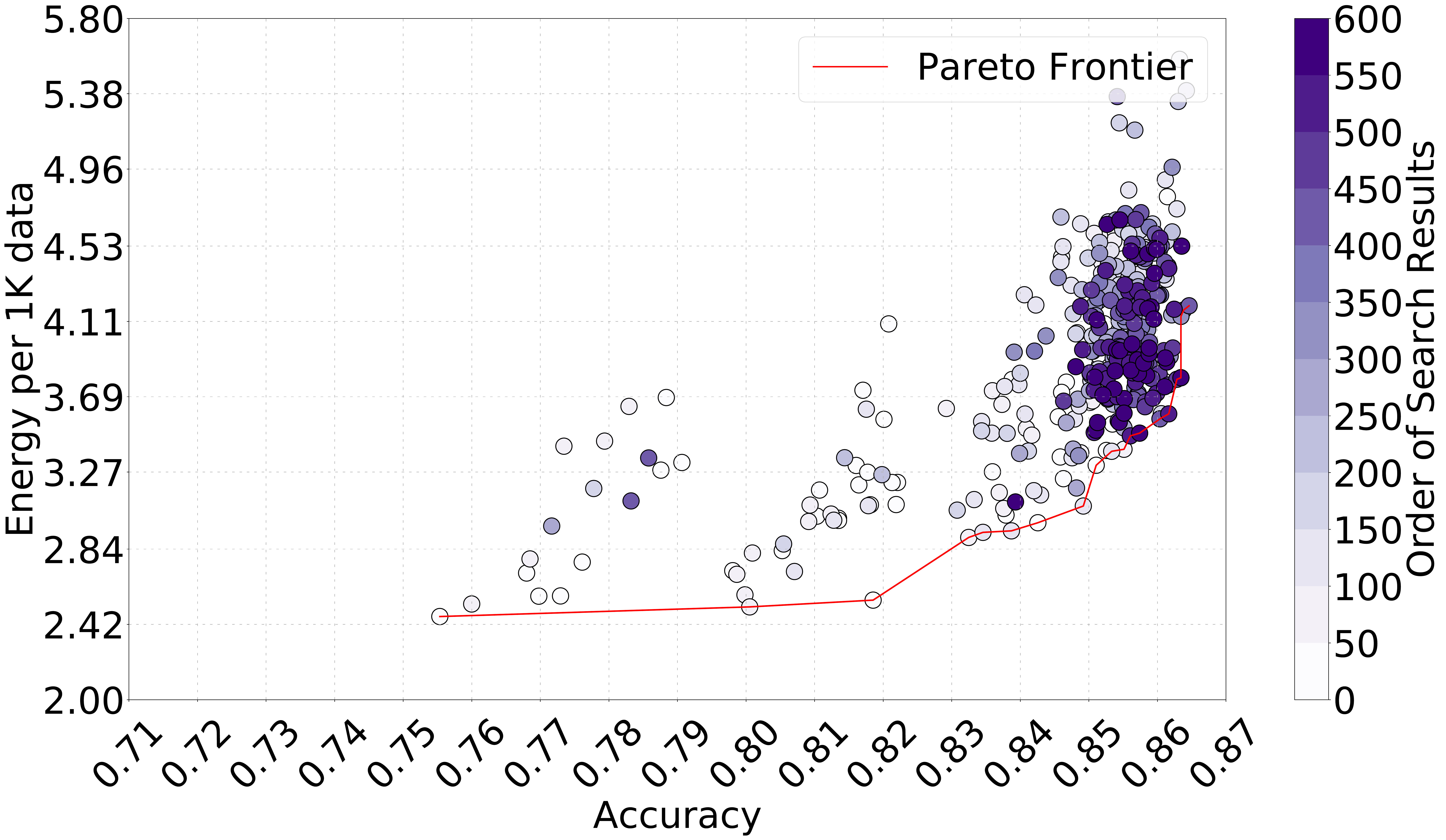}
\label{fig:accu0_85_alex_net}}
\caption{AlexNet Random Search versus Power or Accuracy Constraint}
\label{fig:alexnet_constraint}
\end{figure}

\paragraph{Adaptability.}
\monas allows incorporating application-specific constraints on the search objectives. In Fig.~\ref{fig:70W_alex_net} and Fig.~\ref{fig:accu0_85_alex_net}, we show that our method successfully guided the search process with constraints that either limit the peak power at 70 watts or require a minimum accuracy of $0.85$. In fact, during the 600 iterations of exploration in search space, it can be clearly seen that \monas directs its attention on the region that satisfies the given constraint.  In Fig.~\ref{fig:70W_alex_net}, 340 target networks out of 600 are in the region where peak power is lower than 70 watts.  In Fig.~\ref{fig:accu0_85_alex_net}, 498 target networks out of 600 are in the region where classification accuracy is at least $0.85$. For comparison, we show the exploration process of the Random Search in Fig.~\ref{fig:random_alex_net}, which explored the search space uniformly with no particular emphasis.

In these figures, we also depict the \textit{Pareto Front} of the multi-objective search space (red curves that trace the lower-right boundary of the points). We note that the models that lie on the Pareto Front exhibit the trade-offs between the different objectives and are considered equally good.

  
\begin{figure}[h!]
\centering
\subfigure[70W vs Random]
{\includegraphics[width=0.40\textwidth]{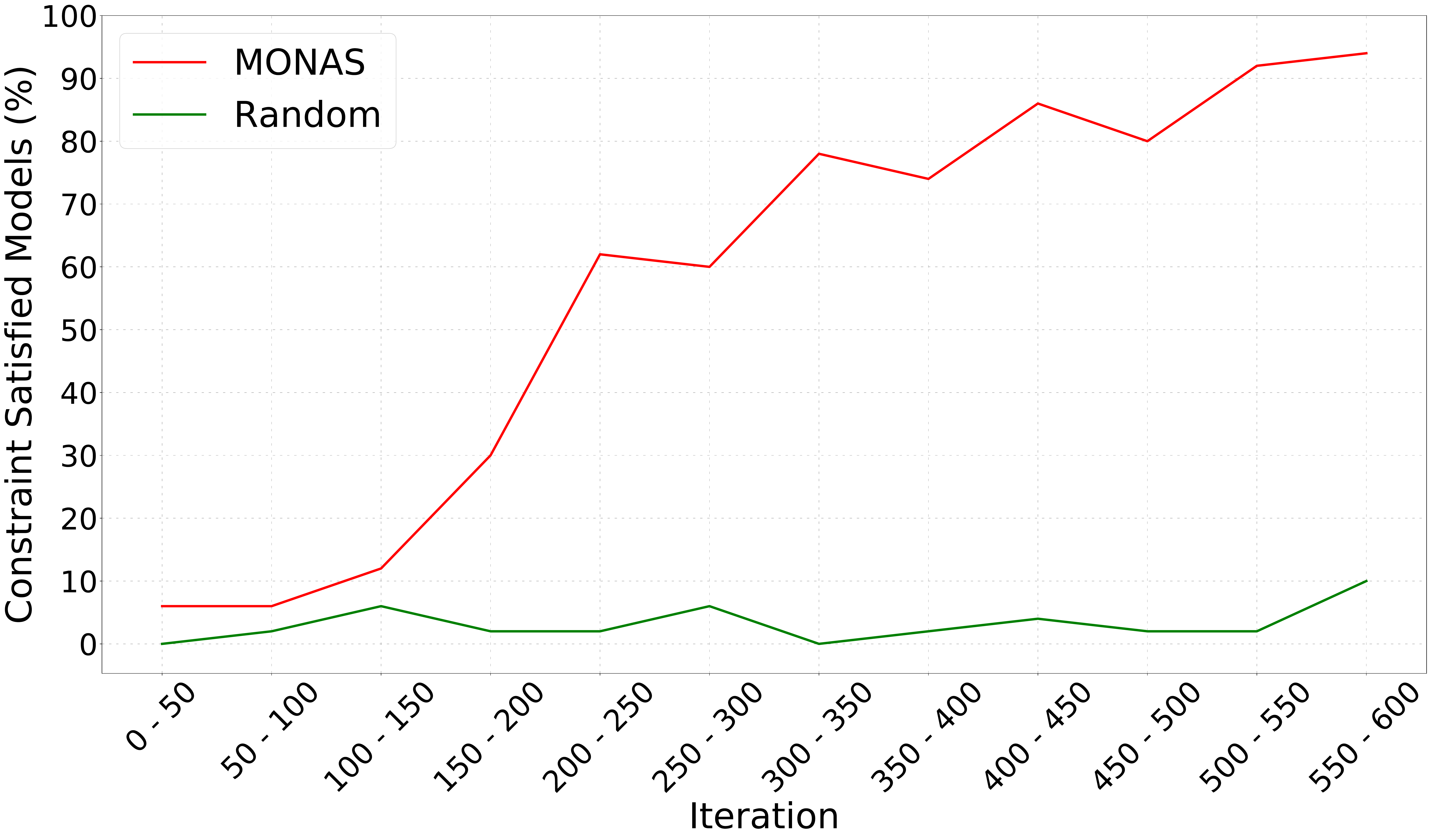}
\label{fig:70vsRan}}
\subfigure[0.85 vs Random]
{\includegraphics[width=0.40\textwidth]{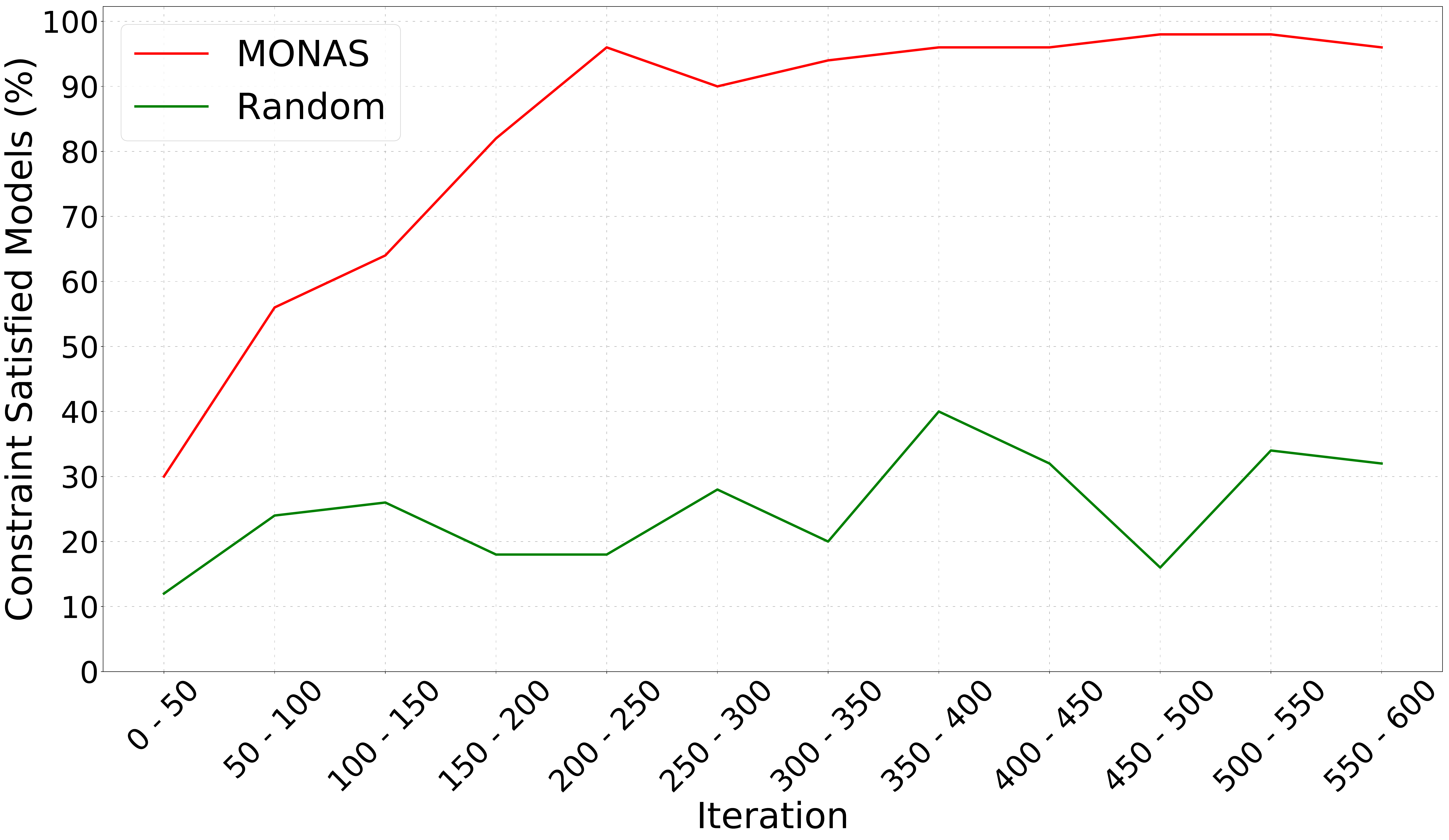}
\label{fig:0.85vsRan}}
\caption{\monas efficiently guides the search toward models satisfying constraints, while the random search has no particular focus.}
\label{fig:effchart}
\end{figure}

\paragraph{Efficiency.}
As shown above, starting with zero knowledge about the search space, \monas can direct the search to the region that satisfies the given constraint.  However, how fast can \monas lock onto the desirable region? In Fig.~\ref{fig:70vsRan}, we show the percentage of architectures that satisfy the constraint of peak power 70Watt at every 50 iterations. After 200 iterations, more than 60\% of architectures generated by \monas satisfy the constraint. Compared to the random search which generates less than 10\% of architectures that satisfy the constraint. Similarly, when given a constraint on classification accuracy, \monas can also guide the search efficiently, as demonstrated in Fig.~\ref{fig:0.85vsRan}.  When the overall search time is limited, \monas is more efficient and has a higher probability to find better architectures and thus outperforms random search.



\begin{figure}[!h]
\centering
\subfigure[$\alpha=0.25$]
{\includegraphics[width=0.45\textwidth]{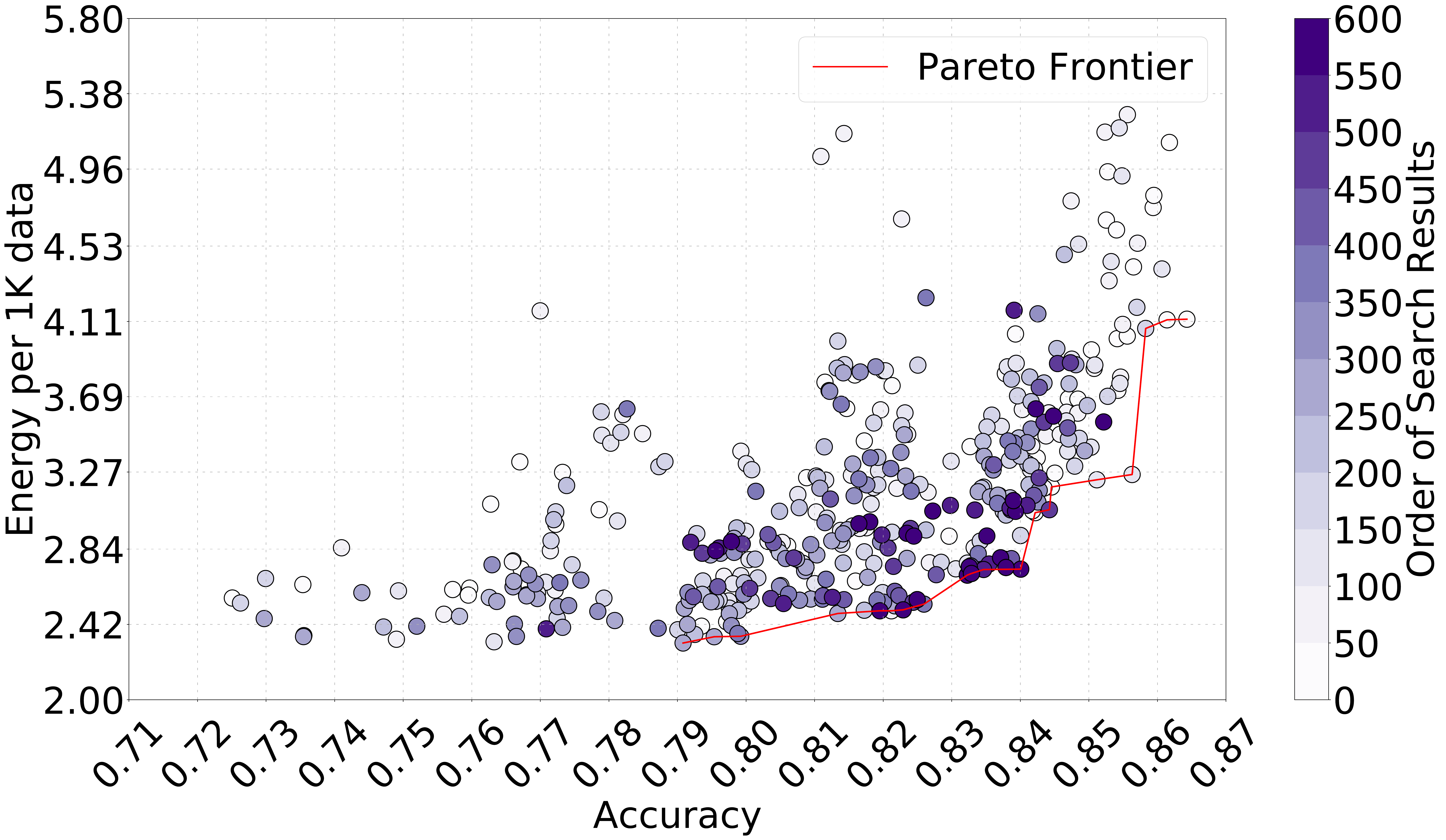}
\label{fig:mix0.25_alex_net}}
\subfigure[$\alpha=0.75$]
{\includegraphics[width=0.45\textwidth]{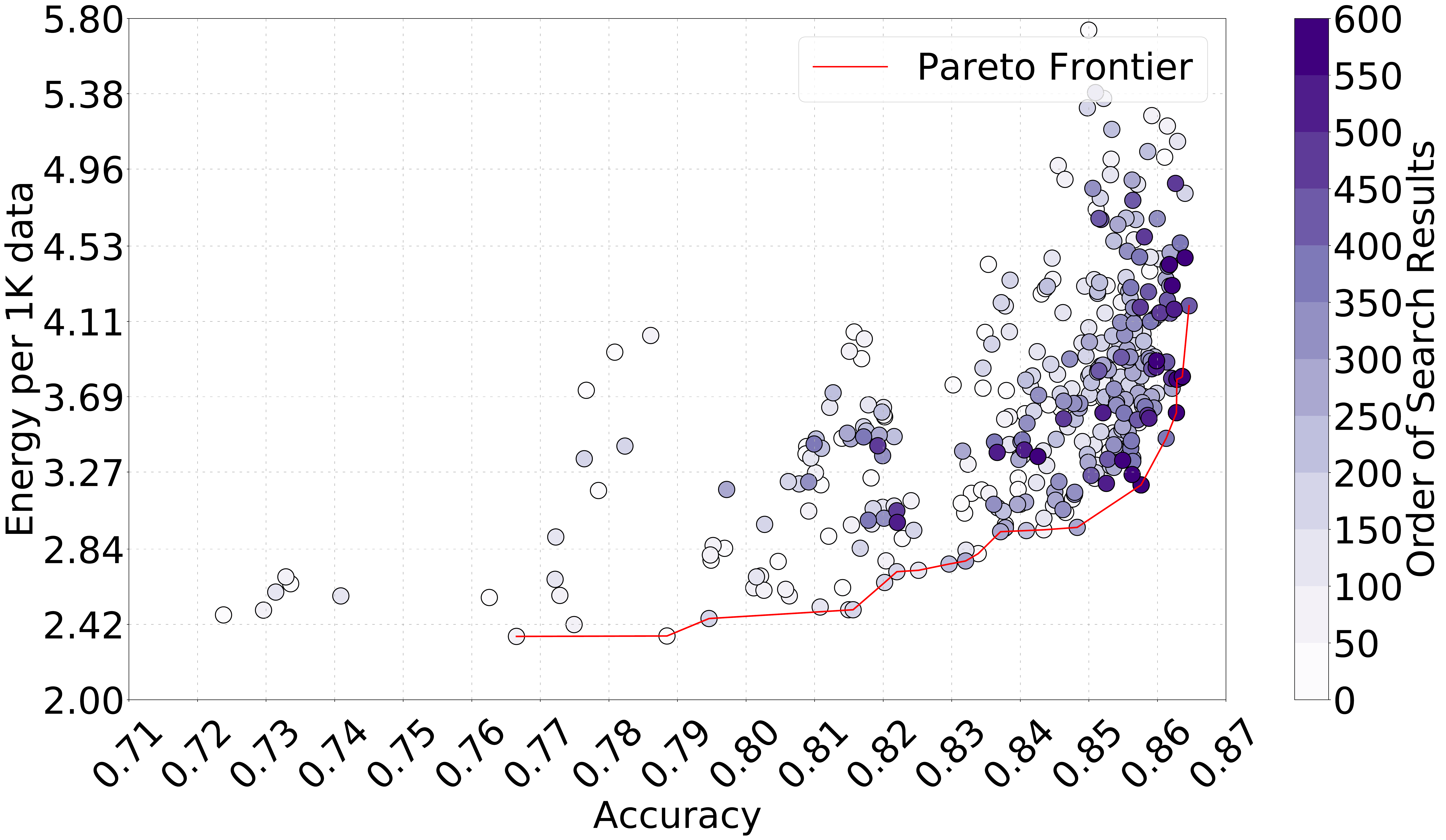}
\label{fig:mix0.75_alex_net}}
\caption{Applying different $\alpha$ when searching AlexNet}
\label{fig:alexnet_mix}
\end{figure}

\begin{figure}[h!]
\centering
\subfigure[Pareto Fronts of searching AlexNet with different $\alpha$.]
{\includegraphics[width=0.45\textwidth]{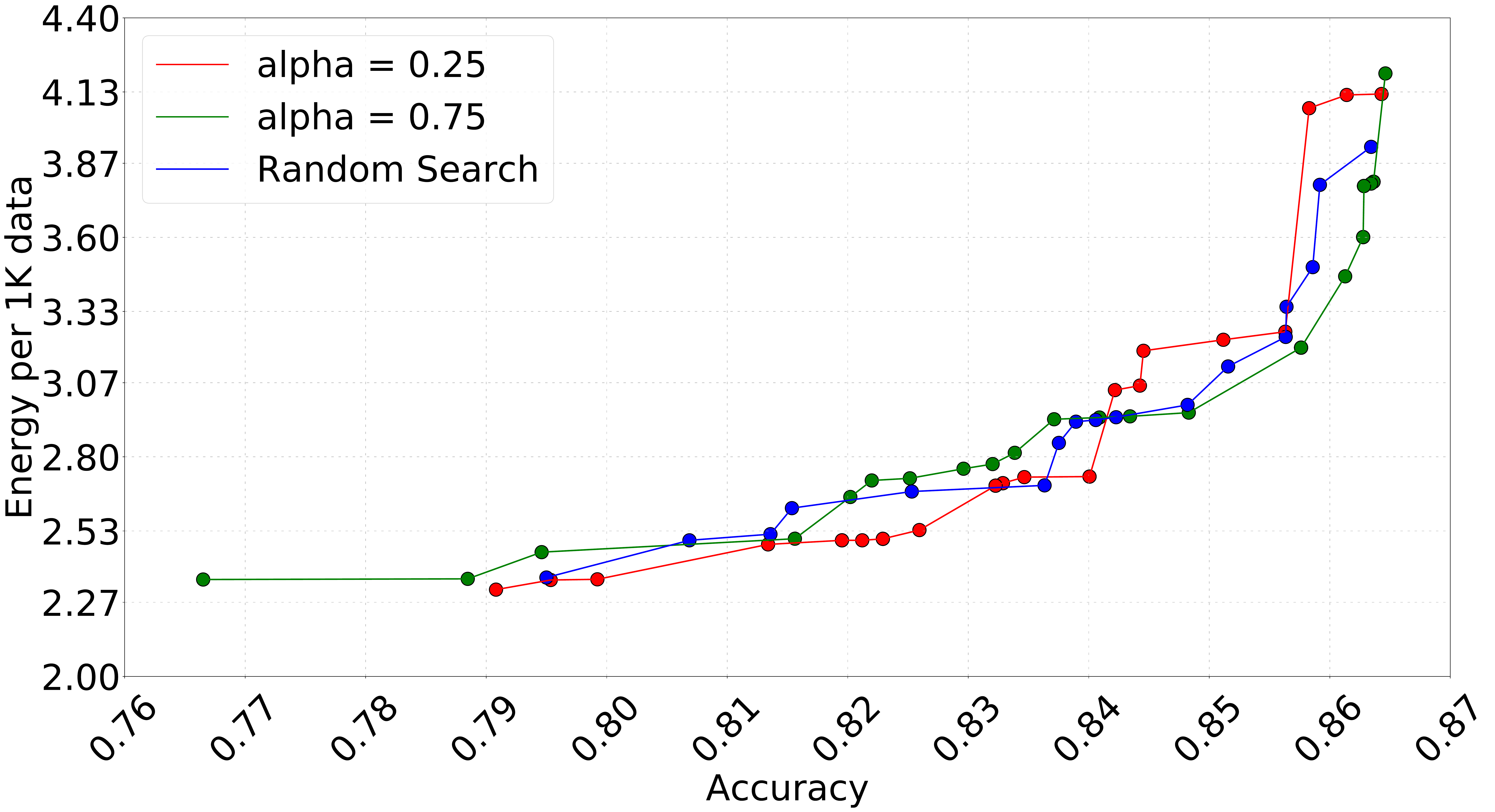}
\label{fig:paretos}}
\subfigure[\monas discovered models that outperform CondenseNet baselines]
{\includegraphics[width=0.38\textwidth]{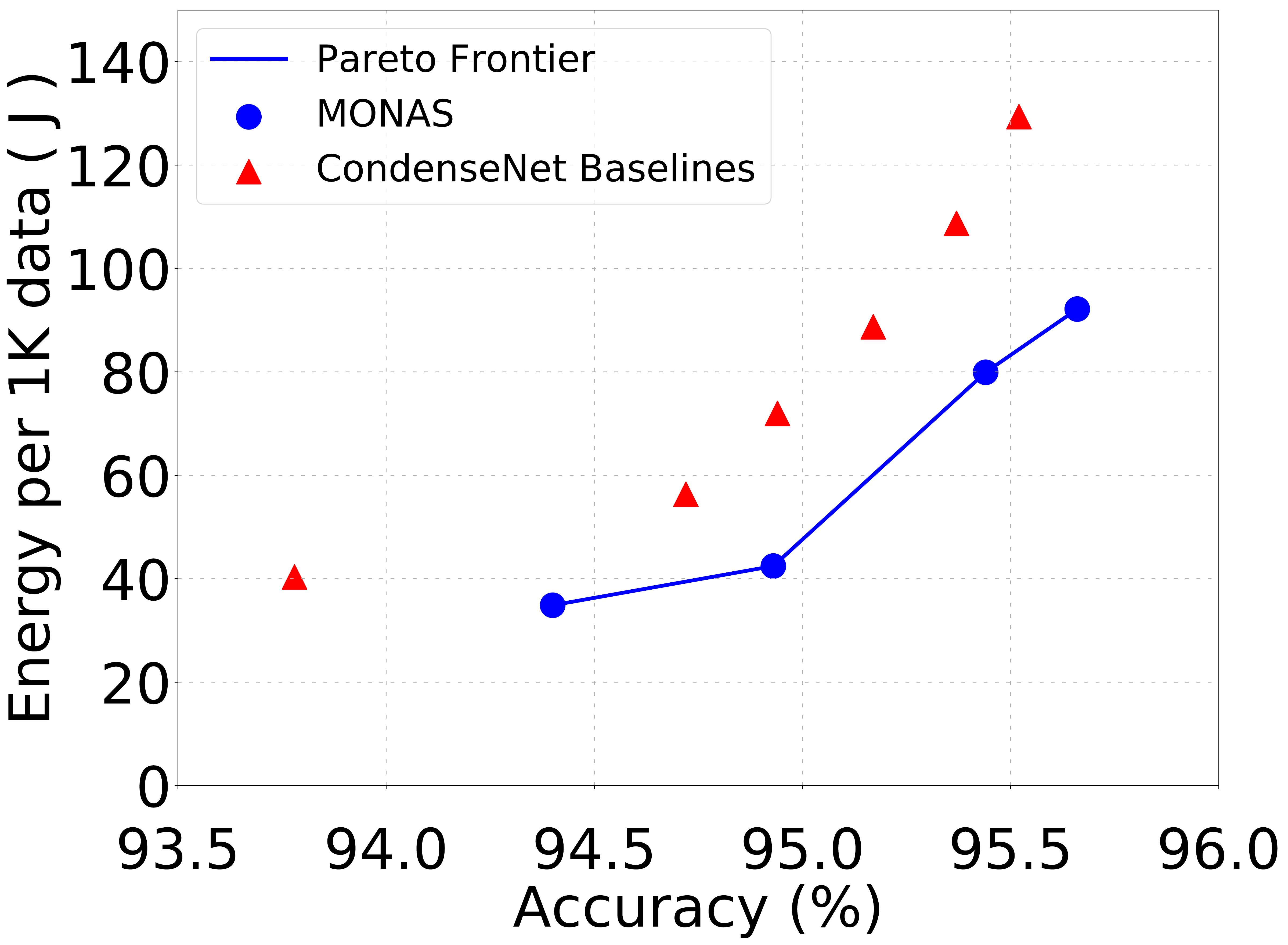}
\label{fig:p_front_condense}
}
\caption{\monas Pareto Fronts on AlexNet and CondenseNet }
\end{figure}

\paragraph{Pareto Frontier.}In Fig.~\ref{fig:mix0.25_alex_net} and Fig.~\ref{fig:mix0.75_alex_net}, we show the search results of 600 architectures when applying different $\alpha$ to the reward function in Eq~\eqref{eq:mix}. \monas demonstrates different search tendency while applying different $\alpha$. To better understand the search tendency, we plot three Pareto Frontiers of $\alpha = 0.25$, $\alpha = 0.75$, and random search, together in Fig~\ref{fig:paretos}. When $\alpha$ is set to 0.25, \monas tends to explore the architectures with lower energy consumption. Similarly, \monas tends to find architectures with higher accuracy when $\alpha$ is set to 0.75. Compared to random search, \monas can find architectures with higher accuracy or lower energy by using $\alpha$ to guide the search tendency.

\paragraph{Discover Better Models.}
In Fig.\ref{fig:p_front_condense}, each point in the figure corresponds to a model. We compare the best CondenseNet models found by \monas and the best ones selected from~\cite{huang2017condensenet}. The Pareto Frontier discovered by \monas demonstrates that \monas has the ability to discover novel architectures with higher accuracy and lower energy than the ones designed by human. Table~\ref{tab:hparams} gives the complete hyperparameter settings and results of the best models selected by \monas.


\subsection{Scalability}
To further justify that the multi-objective reward function works well under a larger search space, we apply our scalable \monass method to find target networks in a 12-layered architecture, similar to the search approach in ENAS \cite{enas}.  Among the 12 layers, the first 4 layers have the largest size, and the size is reduced in half at the next 4 layers, and then further reduced in half at the last 4 layers.  The filters/operations being considered include the convolution 3x3, convolution 5x5, separable convolution 3x3, separable convolution 5x5, average pooling and max pooling. This creates a search space of $1.6 \times 10^{29}$ possible networks.

\begin{figure*}[!h]
\centering
\subfigure[Without MAC Operations Constraint]
{\includegraphics[width=0.40\textwidth]{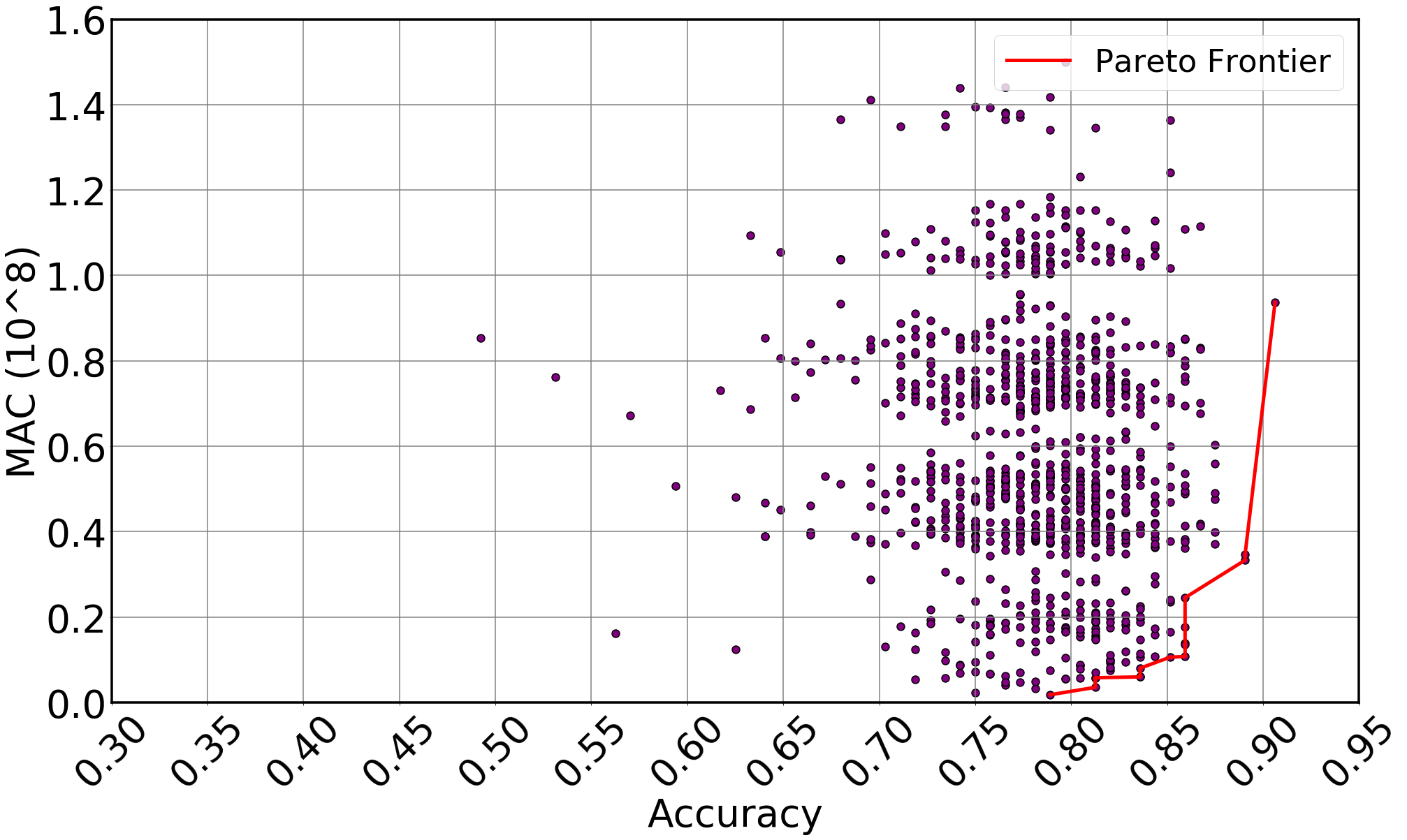}
\label{fig:original_600epoch}}
\subfigure[With MAC Operations Constraint = 0.31]
{\includegraphics[width=0.40\textwidth]{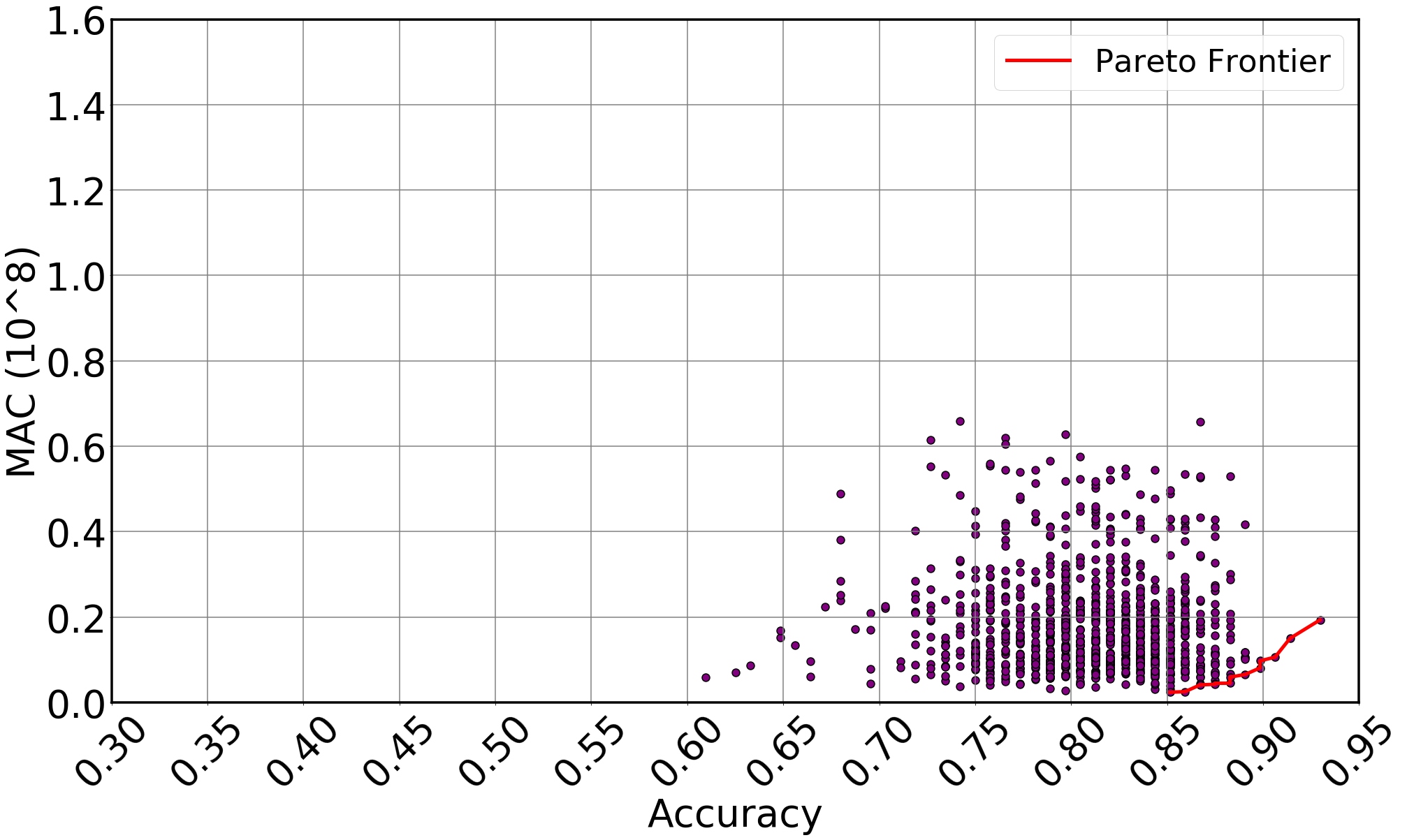}
\label{fig:reward-1_600epoch}}
\caption{The 1000 target networks in (a) are sampled by ENAS and those in (b) are sampled by \monass. Both controllers were trained for 600 epochs.}
\label{fig:moenas_600}
\end{figure*}

In Fig.~\ref{fig:reward-1_600epoch}, we show that \monass successfully guided the search process, even the search space being explored is such larger. As a comparison, we also do a single-objective search, which is actually ENAS~\cite{enas}, using only the validation accuracy (Fig.~\ref{fig:original_600epoch}). In this experiment, both \monass and ENAS train its controller for 600 epochs.  To compare the two trained controllers, we sample 1000 models from each controller, and compare their validation accuracy and the MAC operations required.
It is clear that the target networks found by \monass are biased towards ones that require less MAC and mostly satisfy the given constraint (Fig.~\ref{fig:reward-1_600epoch}). We note that, unlike previous experiments, the accuracy here is the validation accuracy of a mini-batch. 

\begin{figure}[h!]
\centering
{\includegraphics[width=0.45\textwidth]{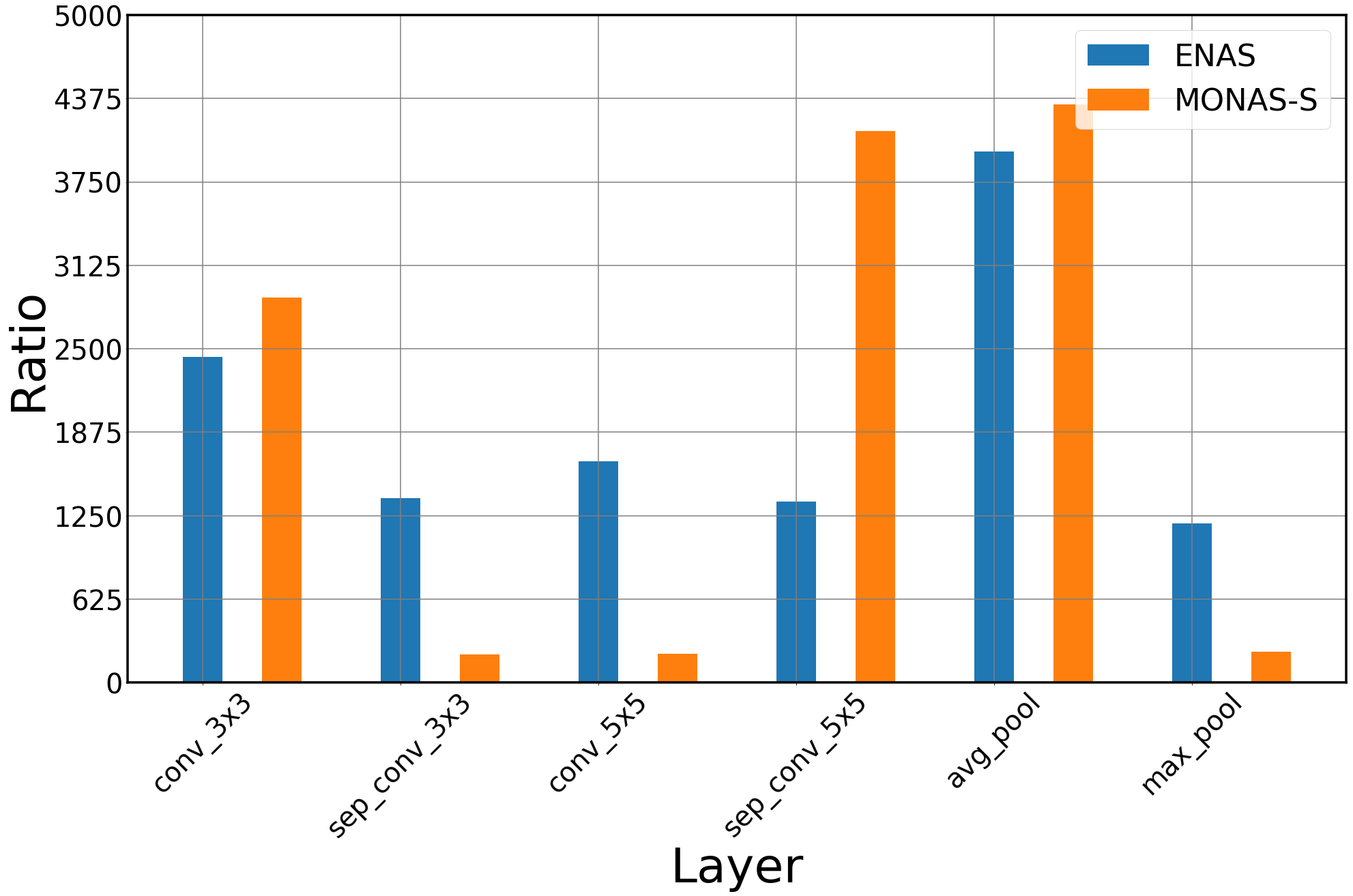}
\caption{Model operation distribution. The blue bars are sampled by ENAS and the orange bars are sampled by \monass.  The Y-axis is the count of the operations being used in a layer of a sampled network.  We sampled 1000 12-layered networks in this experiment, so the sum of either the blue or orange bars is 12000. Model operation distribution.\\\\}
\label{fig:model_operation_distribution}}
\end{figure}

We are interested in knowing which components are preferred by the controller trained by \monass when building (sampling) a model, since such knowledge may give us insights in the design of energy-efficient, high-accuracy models.
In Fig.~\ref{fig:model_operation_distribution}, we show the distribution of the operations selected by the two controllers by ENAS (Fig.~\ref{fig:original_600epoch}) and \monass (Fig.~\ref{fig:reward-1_600epoch}).

Compared to ENAS's controller, \monass's controller tends to select operations with lower MAC operations, while managing to maintain a high-level accuracy at the same time. Operations like depth-wise separable convolution and average pooling are chosen over other operations. This could explain why the target networks found by \monass have lower MAC operations. 

\begin{figure}[ht!]
\centering
{\includegraphics[width=0.50\textwidth]{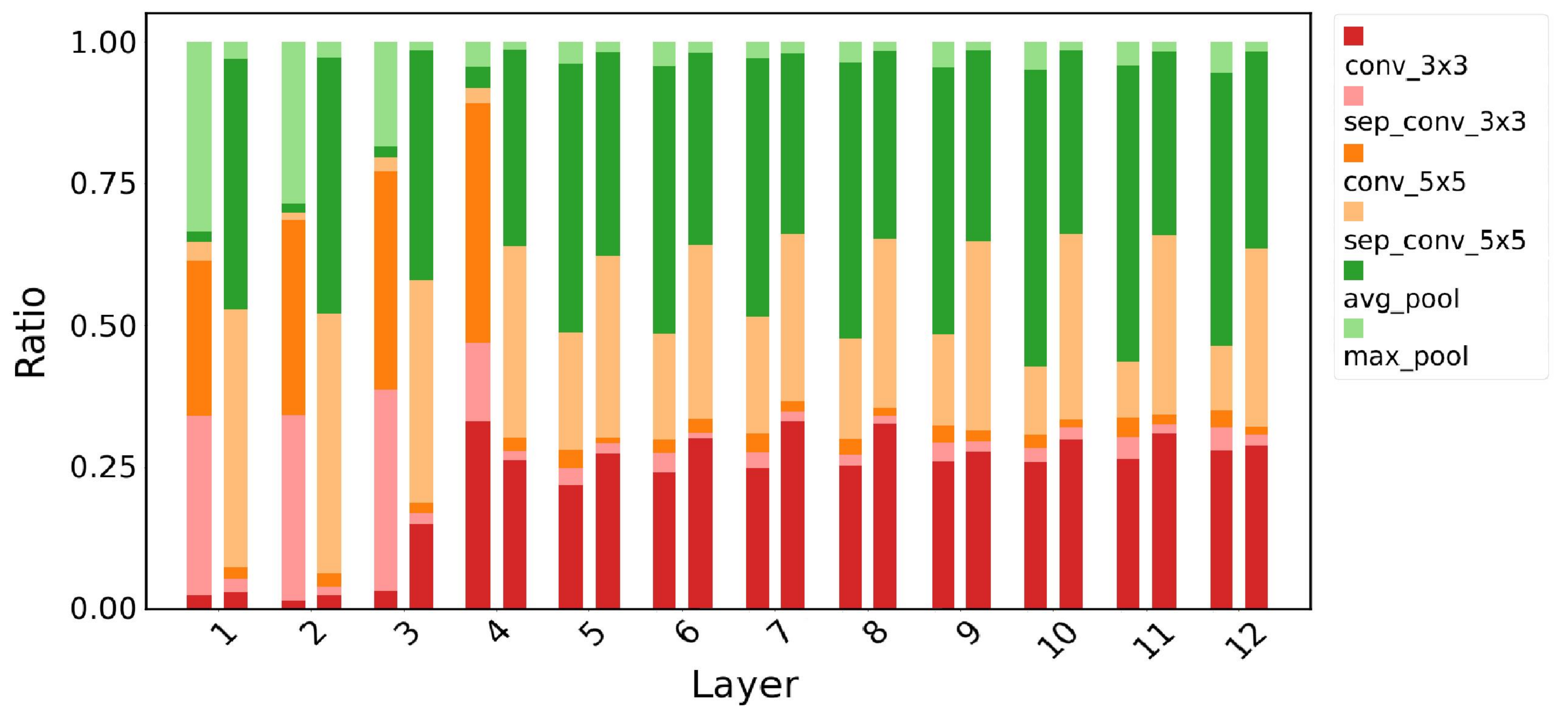}
}
\caption{Layer-wise operation distribution. We show two bars at each layer: the left bar is the distribution of operations sampled by ENAS and the right one is that by \monass. The main difference is at the first 4 layers, within which \monass's controller is more likely to sample MAC-efficient operations such as \texttt{sep\_conv\_5x5}.  We note that the first 4 layers are also the layers with the largest size of feature maps (size is reduced by half every 4 layers in our search space) and tended to need more computation.  This shows that \monas is, surprisingly, able to correctly identify the best opportunity to reduce MAC operations.}
\label{fig:Layer-wise operation distribution}
\end{figure}

Furthermore, as an attempt to understand how these MAC-efficient components are used at each layer, we show the operation distribution at each layer (Fig.~\ref{fig:Layer-wise operation distribution}).  Interestingly, the main differences between target networks sampled by ENAS and those sampled by \monass happened in the first four layers, while the preference of operators are very similar in the later layers.
In the first four layers, \monass prefers MAC-efficient \texttt{sep\_conv\_5x5} and \texttt{avg\_pool} operations, while ENAS prefers \texttt{conv\_5x5} and \texttt{sep\_conv\_3x3}.  We notice that the first four layers are also the ones that have the largest size of feature maps (the size of feature maps reduces by half after every 4 layers) and are tended to need more computation.  It is a pleasant surprise that \monass is able to learn a controller that focuses on reducing MAC at the first 4 layers which have larger feature map sizes and potentially computation-hungry.

\begin{table*}[h!]
  \begin{center}
  \normalsize{
    \begin{tabular}{|c|c|c|c|c|c|c|c|c|} 
    \hline
      \textbf{Model} & \textbf{ST1} & \textbf{ST2} & \textbf{ST3} & \textbf{GR1} & \textbf{GR2} & \textbf{GR3} & \textbf{Error(\%)} & \textbf{Energy(J)} \\

      \hline 
      CondenseNet 122 & 20 & 20 & 20 & 8 & 16 & 32 & 4.48 & 129.37\\
      CondenseNet 110 & 18 & 18 & 18 & 8 & 16 & 32 & 4.63 & 108.74\\
      CondenseNet  98 & 16 & 16 & 16 & 8 & 16 & 32 & 4.83 & 88.73\\
      CondenseNet  86 & 14 & 14 & 14 & 8 & 16 & 32 & 5.06 & 71.98\\
      CondenseNet  74 & 12 & 12 & 12 & 8 & 16 & 32 & 5.28 & 56.35\\
      CondenseNet  50 & 8 & 8 & 8 & 8 & 16 & 32 & 6.22 & 40.35\\
      \hline
      
      \hline 
      \multirow{4}{*}{CondenseNet-\monas}
      & 6 & 14 & 14 & 32 & 32 & 32 & 4.34 & 92.16\\
      & 8 & 8 & 12 & 32 & 32 & 32 & 4.56 & 79.93\\
      & 6 & 12 & 14 & 8 & 32 & 32 & 5.07 & 42.46\\
	  & 14 & 14 & 12 & 4 & 16 & 32 & 5.6 & 34.88 \\
      \hline
      \multicolumn{9}{l}{ST: Stage, GR: Growth, Energy: Energy cost every 1000 inferences}
      \end{tabular}
      }
    \caption{\textbf{\monas models and CondenseNet Baselines}}
    \label{tab:hparams}
  \end{center}
\end{table*}


\begin{figure}[h!]
\centering
\subfigure[$\alpha=0.25$]
{\includegraphics[width=0.45\textwidth]{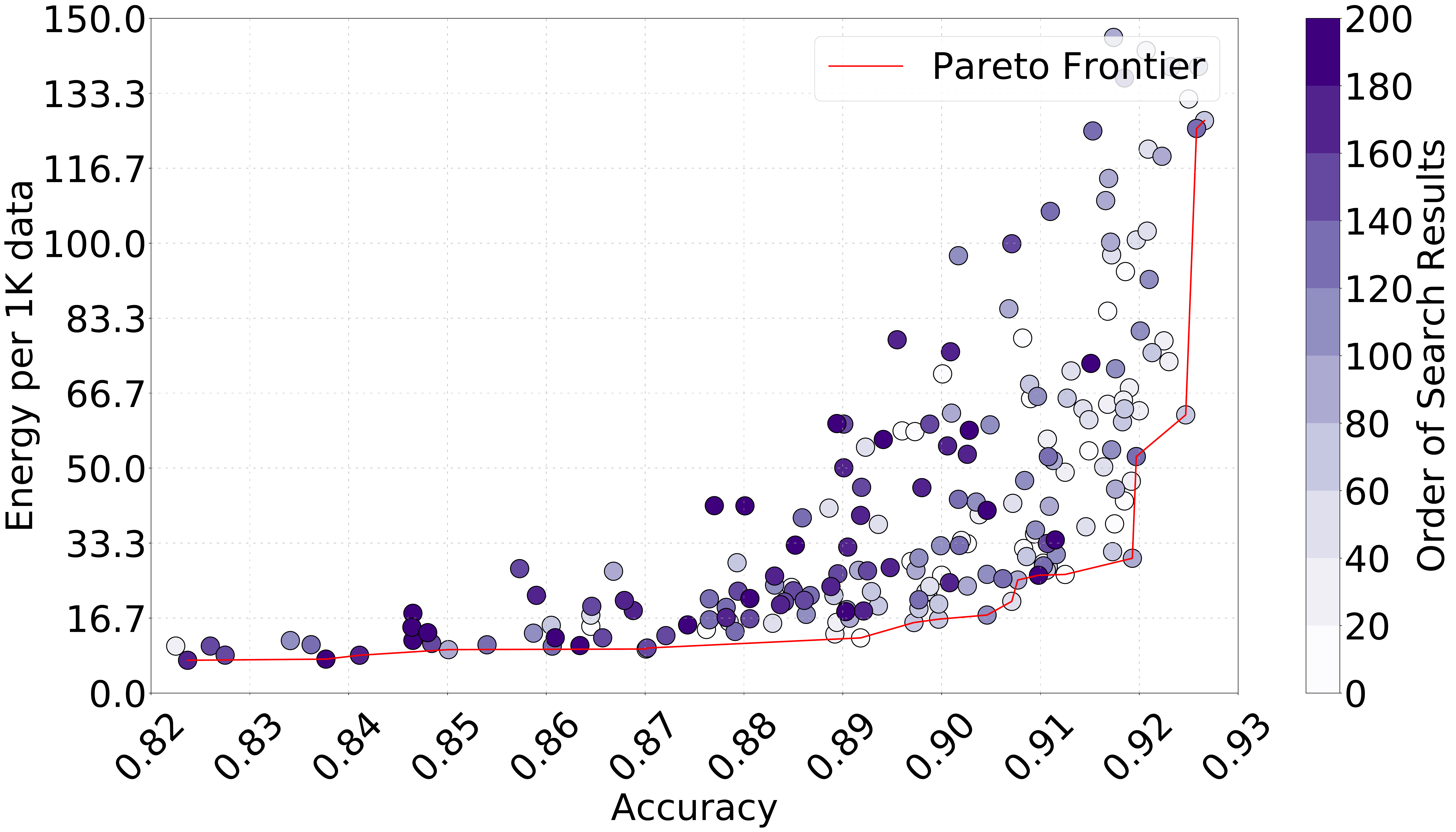}
\label{fig:condensenet_0.25}}
\subfigure[$\alpha=0.75$]
{\includegraphics[width=0.45\textwidth]{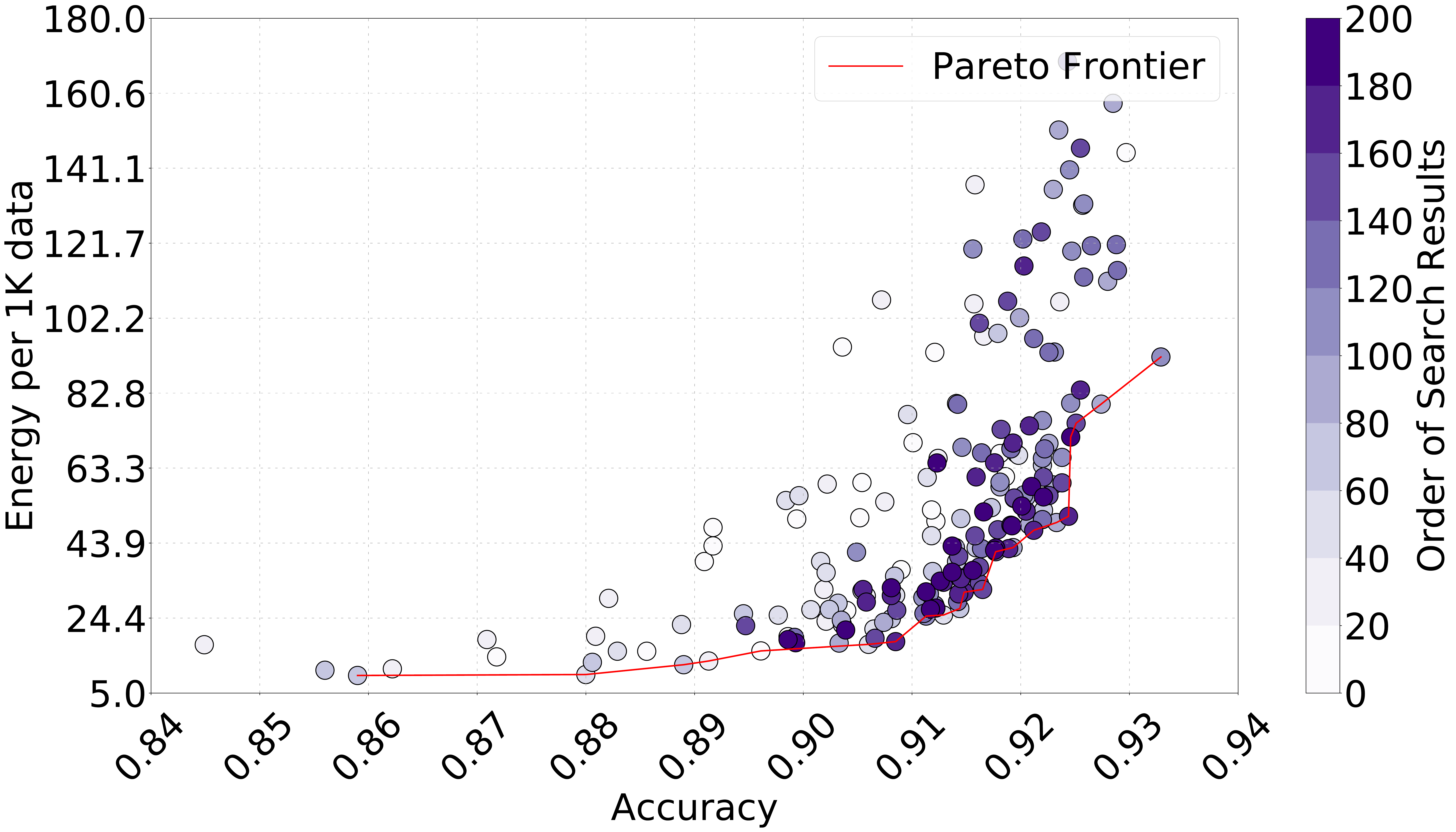}
\label{fig:condensenet_0.75}}
\caption{Applying different $\alpha$ when searching CondenseNet}
\label{fig:condensenet_mix}
\end{figure}

\subsection{Training Details for Target Networks}
\paragraph{AlexNet:}
We design our \rn with one-layer LSTM with 24 hidden units and train with ADAM optimizer \cite{adam2015} with learning rate = 0.03. The \tn is constructed and trained for 256 epochs after the \rn selects the hyperparameters for it. Other model settings refer to the TensorFlow tutorial.\footnote{www.tensorflow.org\//tutorials\//deep\_cnn} 
\paragraph{CondenseNet:}
Our \rn is constructed with one-layer LSTM with 20 hidden units and trained with ADAM optimizer with learning rate = 0.008. We train  the \tn for 30 epochs and other configurations refer to the \baselineB \cite{huang2017condensenet} public released code.\footnote{github.com\//ShichenLiu\//CondenseNet}
We train the best performing models found by \monas for 300 epochs on the whole training set and report the test error.
\paragraph{\monass:}
We build \monass controller by extending ENAS with our \monas framework. We train the sharing weight and controller for 600 epochs. For the other configurations we refer to the ENAS public release code.\footnote{github.com\//melodyguan\//enas}



\section{Conclusion}
\label{sec:conclusion}
In this paper, we propose \monas, a multi-objective architecture search framework based on deep reinforcement learning. We show that \monas can adapt to application-specific constraints and effectively guide the search process to the region of interest.
In particular, when applied on \baselineB, \monas discovered models that outperform the ones reported in the original paper, with higher accuracy and lower power consumption.  In order to work with larger search spaces and reduce search time, we extend the concepts of \monas and propose the \monass, which is scalable and fast. \monass explores larger search space in a shorter amount of time.




\newpage
\bibliographystyle{aaai}
\bibliography{06Reference.bib}






\end{document}